\documentclass[lettersize,journal]{IEEEtran}
\usepackage{amsmath,amsfonts}
\usepackage{algorithmic}
\usepackage{algorithm}
\usepackage{array}
\usepackage[caption=false,font=normalsize,labelfont=sf,textfont=sf]{subfig}
\usepackage{textcomp}
\usepackage{stfloats}
\usepackage{url}
\usepackage{verbatim}
\usepackage{graphicx}
\usepackage{cite}
\usepackage{hyperref}
\begin{document}

\title{\fontsize{24}{28}\selectfont Antenna Near-Field Reconstruction from Far-Field Data Using Convolutional Neural Networks}


\author{
	\IEEEauthorblockN{Sahar Bagherkhani\IEEEauthorrefmark{1}, 
Jackson Christopher Earls\IEEEauthorrefmark{2}, Franco De Flaviis\IEEEauthorrefmark{1}, Pierre Baldi\IEEEauthorrefmark{2}}
	
\IEEEauthorblockA{\IEEEauthorrefmark{1} Department of Electrical Engineering and Computer Science, University of California at Irvine, Irvine, USA \\}

\IEEEauthorblockA{\IEEEauthorrefmark{2} Department of Computer Science, 
	University of California at Irvine, Irvine, USA \\
	(sbagherk@uci.edu, jearls@uci.edu, franco@uci.edu, pfbaldi@ics.uci.edu)}
    
	}



\pagenumbering{gobble}

\maketitle

\begin{abstract}
Electromagnetic field reconstruction is crucial in many applications, including antenna diagnostics, electromagnetic interference analysis, and system modeling. This paper presents a deep learning-based approach for Far-Field to Near-Field (FF-NF) transformation using Convolutional Neural Networks (CNNs). The goal is to reconstruct near-field distributions from the far-field data of an antenna without relying on explicit analytical transformations. The CNNs are trained on paired far-field and near-field data and evaluated using mean squared error (MSE). The best model achieves a training error of 0.0199 and a test error of 0.3898. Moreover, visual comparisons between the predicted and true near-field distributions demonstrate the model’s effectiveness in capturing complex electromagnetic field behavior, highlighting the potential of deep learning in electromagnetic field reconstruction.
\end{abstract}


\section{Introduction}

Accurate antenna characterization and field measurements play a crucial role in electromagnetic (EM) design, as they help to ensure optimal performance in wireless communication, radar, and satellite systems \cite{balanis2016antenna,10401876}. To achieve this, two primary measurement techniques are used to evaluate antennas: near-field (NF) and far-field (FF) measurements. NF measurements provide detailed spatial field distributions but often require complex experimental setups. In contrast, FF measurements are commonly used for antenna pattern assessment, as they offer a simplified representation of the EM field at large distances \cite{balanis2016antenna}. However, direct FF measurements require a sufficiently large test range—typically several wavelengths—which presents challenges for electrically large antennas.

To overcome these limitations, NF measurement techniques have been widely developed, where the EM field is sampled in the near-field region and mathematically transformed to the far-field using analytical methods such as modal expansions and Fourier transforms. Near-Field to Far-Field (NF-FF) transformations have been extensively studied and implemented in planar, cylindrical, and spherical scanning configurations \cite{1143727}. In contrast, the Far-Field to Near-Field (FF-NF) transformation is a more challenging inverse problem that reconstructs the near-field distribution, which represents a more complete form of the electromagnetic field, from far-field data. This transformation is particularly valuable in applications where direct NF measurement is not feasible, such as antenna diagnostics, electromagnetic interference (EMI) analysis, and computational EM modeling. The primary challenge in FF-NF transformation arises from the fact that far-field data contains only partial information about the original electromagnetic field. In particular, radial and evanescent wave components, which are present in the near field, do not propagate to the far field and are therefore not directly available for reconstruction. As a result, reconstructing the near-field distribution from far-field measurements is challenging and requires advanced mathematical techniques to estimate the missing field components based on the available far-field data.

Machine learning (ML) has proven to be an effective approach for solving complex inverse problems \cite{9211490, baldi2021deep}. Due to the complexity and incomplete nature of far-field data, data-driven methods such as machine learning may offer a promising solution for estimating the near-field components. In this paper, we propose a Convolutional Neural Network (CNN)-based approach for FF-NF transformation, using deep learning to reconstruct near-field distributions from far-field measurements. By training CNNs on a large dataset of simulated FF and corresponding NF component pairs, the network learns to reconstruct missing near-field information.

The remainder of this paper is organized as follows: Section II describes the dataset generation process and the architecture of the best-performing CNN. Section III presents the numerical results and validation. Finally, Section IV concludes the paper.

\section{Far-Field to Near-Field Transformation}
Electromagnetic fields around an antenna are divided into two primary regions: the near-field region and the far-field region, each exhibiting distinct field characteristics. The boundaries between these regions are defined based on the distance from the antenna. The near-field region is the area closest to the antenna, where evanescent waves and reactive components dominate. It is further divided into the reactive near-field, where energy is primarily stored rather than radiated, and the radiating near-field (Fresnel region), where radiation begins to dominate but the field distribution remains dependent on distance \cite{482029}. Near-field measurements are commonly performed within this region.

At greater distances, typically defined as $R \geq \frac{2D^2}{\lambda}$  (where $D$ is the antenna's largest dimension and $\lambda$ is the wavelength), the far-field region begins, where electromagnetic waves propagate as plane waves with a fixed angular distribution \cite{balanis2016antenna}. While there are no abrupt boundaries between these regions, their distinction is crucial in applications such as FF-NF transformation, where missing near-field details must be reconstructed from far-field measurements.

The goal of this paper is to evaluate whether a machine learning approach can effectively perform the far-field to near-field transformation while keeping dataset generation computationally feasible.

\subsection{Electromagnetic Modeling and Dataset Generation}
One of the main challenges in applying machine learning to electromagnetic problems is generating a sufficiently large dataset. EM simulations are computationally expensive and require significant time and resources.

\begin{figure}[!t]
\centering
\includegraphics[width=\linewidth]{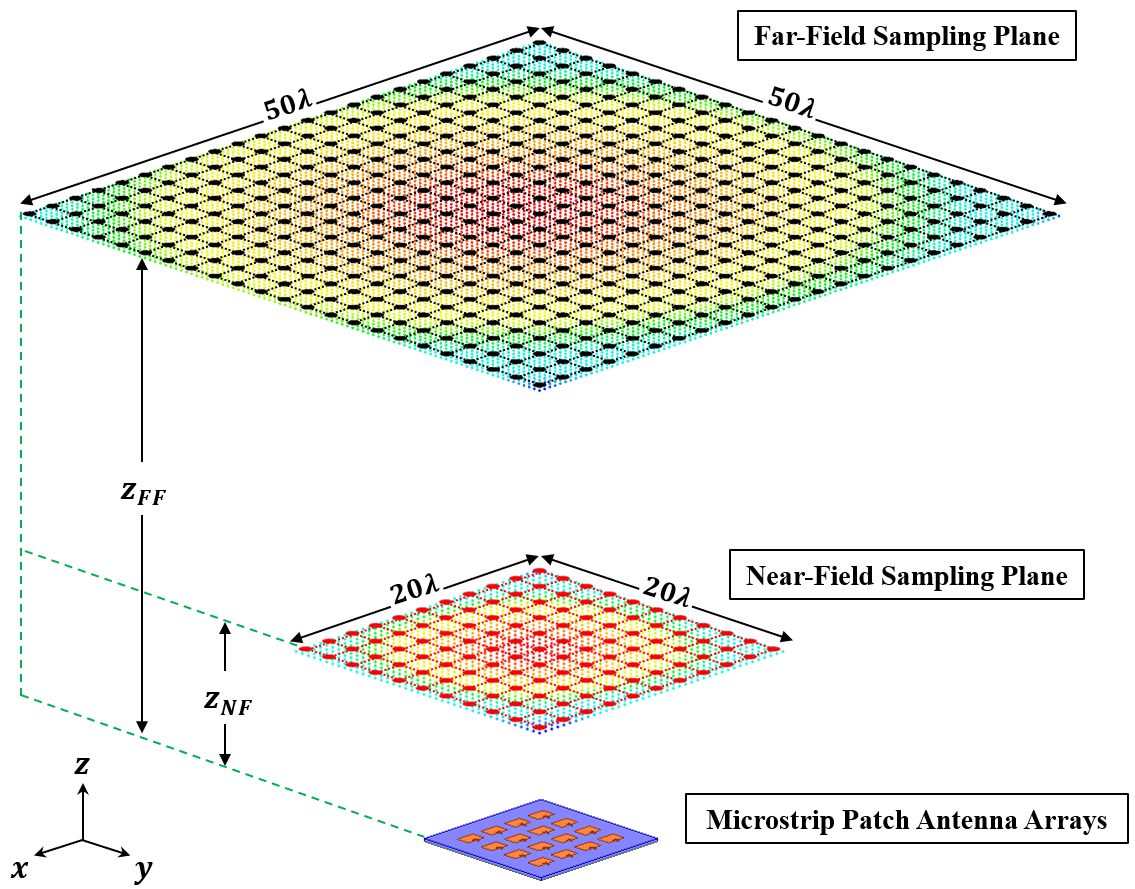}
\caption{Antenna array with near-field and far-field sampling grids used for dataset generation ($Z_{\text{NF}}=4.5\lambda$, $Z_{\text{FF}}=25\lambda$).}
\label{Fig_1}
\end{figure}

To create a large dataset, we consider a $4\times4$ microstrip patch antenna array operating at 30 GHz. For this problem, dataset generation requires both near-field and far-field data of the antenna. One approach is to solve the antenna in a full-wave simulator once and then, by varying the phase excitations of the antenna elements in postprocessing, generate different near-field and far-field distributions. Fig.~\ref{Fig_1} shows the antenna array and the near-field and far-field sampling grids used for dataset generation.

The near-field region is defined at a distance of $Z_{\text{NF}}=4.5\lambda$ on a grid of $20\lambda \times 20\lambda$ with a sampling resolution of $\lambda/2$. Additionally, the far-field region is defined at $Z_{\text{FF}}=25\lambda$ on a grid of $50\lambda \times 50\lambda$, also with a sampling resolution of $\lambda/2$. The simulation is performed using Ansys HFSS, with a Python-HFSS user interface developed to automate the dataset generation process. Each of the 16 antenna elements can have a phase value of $0^\circ, 90^\circ, 180^\circ,$ or $270^\circ$, resulting in a total of $4^{16}$ possible phase configurations.

The Python script randomly assigns phase values to the antenna elements and sends these configurations to Ansys HFSS for a predefined antenna model. After postprocessing, the magnitude and phase of the electric field components—$E_x, E_y, E_z$—in both the near-field and far-field regions are extracted to build the dataset. The dataset generated from these simulations provides the necessary training data for evaluating the machine learning model’s ability to reconstruct near-field information from far-field measurements.

The dataset consists of both far-field and near-field data. The dimensions of the far-field and near-field distributions are $6\times101\times101$ and $6\times41\times41$, respectively. The first dimension represents the real and imaginary parts of the electric field components $E_x, E_y$, and $E_z$. The second and third dimensions correspond to the grid width and height.

\subsection{Architecture of the Convolutional Neural Network}
To determine the optimal model architecture and hyperparameters for the convolutional neural network, we employed SHERPA \cite{hertel2020sherpa}, a hyperparameter optimization library. We conducted a grid search for each tunable hyperparameter. The optimized CNN includes a convolutional layer, followed by a ReLU activation function, an average pooling layer, and a linear output layer.
\subsubsection{Convolutional Layer}
The convolutional layer is composed of 6 input channels, 10 output channels, a kernel size of 5, and a kernel stride of 1. These input channels correspond to the 6 complex components of the input far-field signal. We tested output channel values ranging from 6 to 24 and found the optimal number to be 10. After evaluating all combinations of kernel sizes (3 and 5) and strides (1 and 2), we found that a kernel size of 5 and a stride of 1 were optimal.
\subsubsection{ReLU Activation Function}
The ReLU activation function is commonly used in convolutional neural networks due to its computational efficiency and ability to mitigate vanishing gradients. In our model, we experimented with ReLU, Tanh, and Leaky ReLU activation functions. Among these, the ReLU activation function was found to be the most computationally efficient and accurate.
\subsubsection{Average Pooling Layer} 
We experimented with both max pooling and average pooling layers in our model. Average pooling was found to be more effective, resulting in a significant improvement in accuracy. We tested kernel values of 2 and 3 for the pooling layer and found that 2 was optimal.
\subsubsection{Linear Layer}
Prior to selecting a single linear output layer, we tested several model variations, including architectures with multiple fully connected layers. Specifically, we evaluated designs with two and three fully connected layers (including the output layer), using different activation functions between the layers. A single linear output layer proved to be the optimal architecture. The linear layer has an input dimension corresponding to the product of the height, width, and number of output channels following the average pooling layer. Finally, the output dimension of the linear layer is 10,086, which corresponds to the dimension of $6\times41\times41$ of the predicted near-field signal.
\subsubsection{Model Training}
The CNN was trained on 4,050 data points and tested on 450. Training was conducted over 700 epochs with a batch size of 32, using the Adam optimizer and a learning rate of $5 \times 10^{-4}$, along with a linear learning rate scheduler. The scheduler had a start factor of 1.0 and an end factor of 0.005, applied over 300 iterations. We explored learning rates ranging from $10^{-3}$ to $10^{-6}$  and employed a learning rate scheduler to reduce training time and enhance model accuracy. Additionally, we experimented with batch sizes between 8 and 64 before determining that 32 was optimal.
\section{Results}
\begin{figure}[t]
    \centering
    \includegraphics[width=\linewidth]{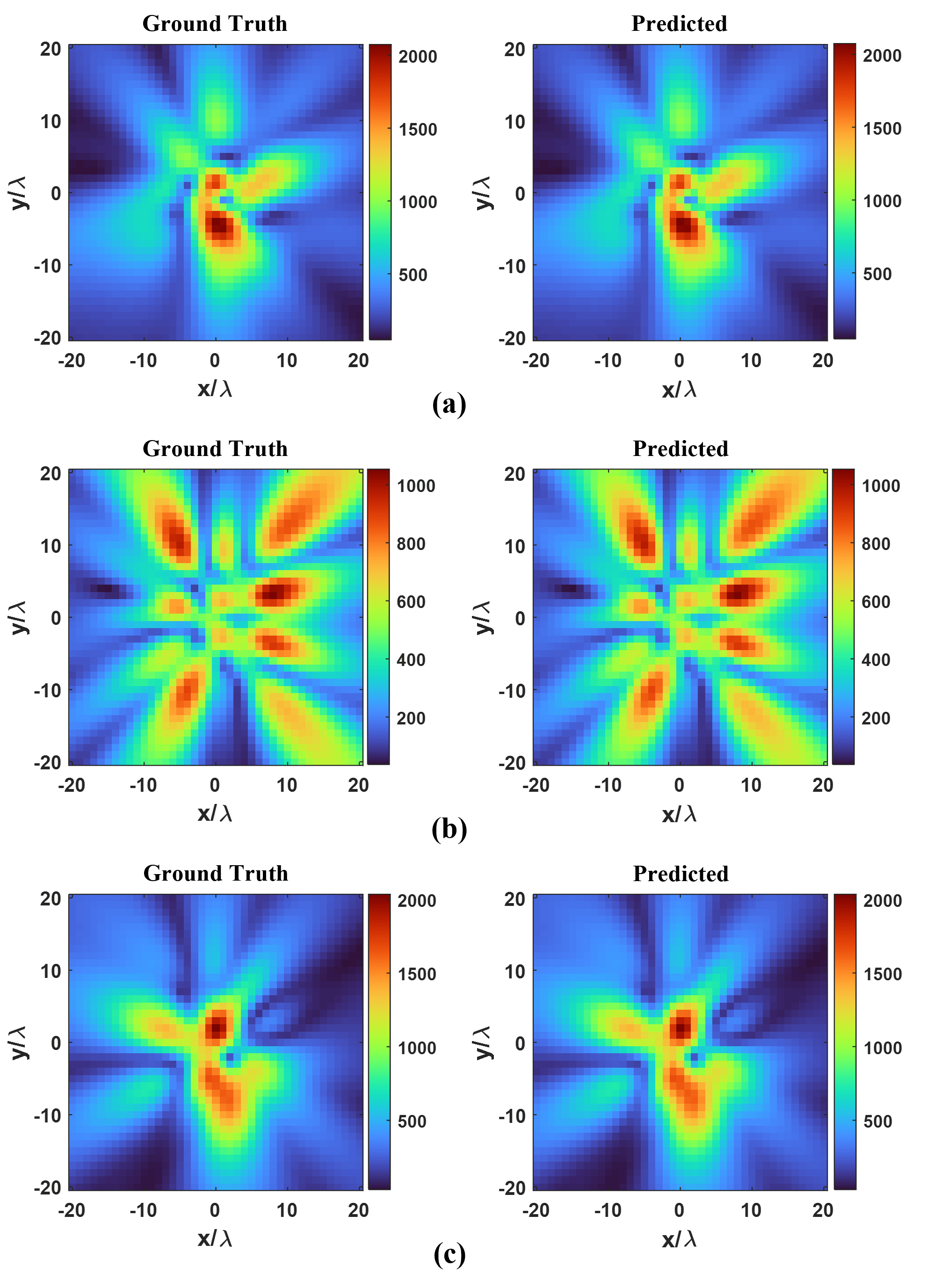}
    \caption{Comparison of predicted and ground truth total near-field magnitude distributions for three representative examples. Each row corresponds to one example, with the ground truth shown in the left column and the CNN-predicted field in the right.}
    \label{Fig_2}
\end{figure}

\begin{figure}[t]
    \centering
    \includegraphics[width=\linewidth]{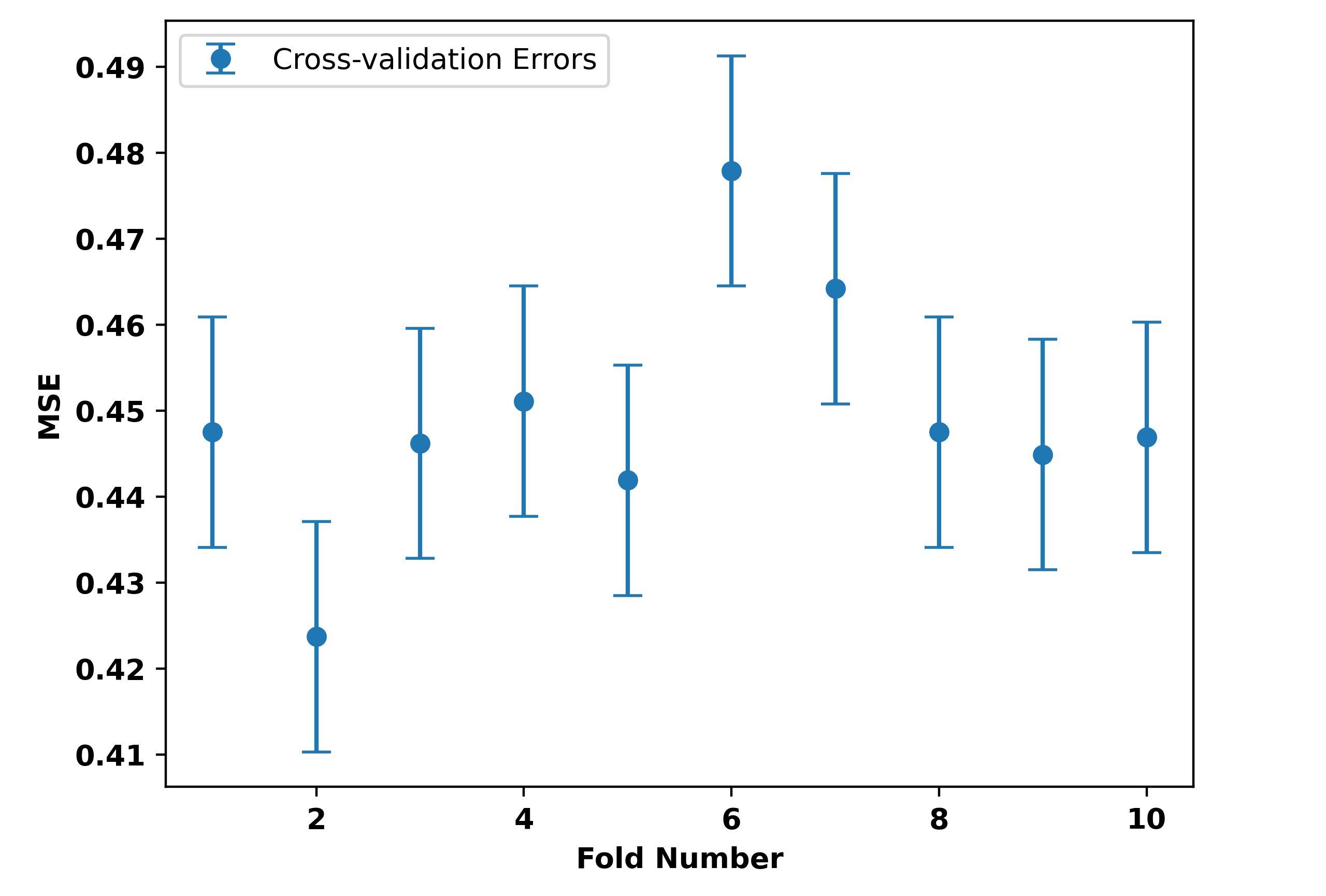}
    \caption{Error bar resulting from 10-fold cross-validation.}
    \label{err}
\end{figure}

In this section, we demonstrate the performance of the optimized CNN model in reconstructing the near-field distribution from far-field measurements. Fig.~\ref{Fig_2} presents side-by-side comparisons of the complex magnitude of the total electric field for the predicted near-field distributions and the corresponding ground truth across three representative examples. The optimized CNN model achieved a training error of 0.0199 mean squared error (MSE) and a test error of 0.3898 MSE. The near-field dataset contains values ranging from -3,748.556 to 3,456.257. Given this range, the optimized CNN model attains a normalized root mean squared error (NRMSE) of $8.66 \times 10^{-5}$ on the test set. To further validate the ability of the CNN model to generalize to new data, we performed 10-fold cross-validation. The resulting mean and standard deviation from cross-validation are $\mu_{MSE}=0.449$ and $\sigma_{MSE}=0.0141$. Fig.~\ref{err} illustrates the error bars obtained from the cross-validation. An additional advantage of using CNNs for near-field reconstruction is their computational efficiency. The optimized model can generate a near-field distribution from a far-field input in under one minute. These results demonstrate that CNNs can effectively reconstruct near-field distributions from far-field data, accurately capturing the spatial characteristics of the electromagnetic field while maintaining time efficiency. The visual agreement between predicted and true field patterns further confirms the model’s ability to generalize to unseen data.

\section{Conclusion}
This paper proposed a Convolutional Neural Network (CNN)-based approach for Far-Field to Near-Field transformation, using deep learning to estimate near-field distributions from the far-field data of an antenna. A large dataset was generated through Ansys HFSS simulations of a $4\times4$ microstrip patch antenna array at 30 GHz, with varying phase excitations to train the model. The results demonstrate that the CNN successfully reconstructs near-field components, providing accurate field estimations based on far-field measurements.

\bibliographystyle{IEEEtran}
\bibliography{ref}

\begin{thebibliography}{1}
\providecommand{\url}[1]{#1}
\csname url@samestyle\endcsname
\providecommand{\newblock}{\relax}
\providecommand{\bibinfo}[2]{#2}
\providecommand{\BIBentrySTDinterwordspacing}{\spaceskip=0pt\relax}
\providecommand{\BIBentryALTinterwordstretchfactor}{4}
\providecommand{\BIBentryALTinterwordspacing}{\spaceskip=\fontdimen2\font plus
\BIBentryALTinterwordstretchfactor\fontdimen3\font minus \fontdimen4\font\relax}
\providecommand{\BIBforeignlanguage}[2]{{%
\expandafter\ifx\csname l@#1\endcsname\relax
\typeout{** WARNING: IEEEtran.bst: No hyphenation pattern has been}%
\typeout{** loaded for the language `#1'. Using the pattern for}%
\typeout{** the default language instead.}%
\else
\language=\csname l@#1\endcsname
\fi
#2}}
\providecommand{\BIBdecl}{\relax}
\BIBdecl

\bibitem{balanis2016antenna}
C.~A. Balanis, \emph{Antenna theory: analysis and design}.\hskip 1em plus 0.5em minus 0.4em\relax John wiley \& sons, 2016.

\bibitem{10401876}
S.~Alamdar, K.~Mohammadpour-Aghdam, and H.~Khalili, ``A compact monopulse antenna array with suppressed mutual coupling using broadband schiffman phase shifters,'' \emph{IEEE Access}, vol.~12, pp. 11\,936--11\,944, 2024.

\bibitem{1143727}
A.~Yaghjian, ``An overview of near-field antenna measurements,'' \emph{IEEE Transactions on Antennas and Propagation}, vol.~34, no.~1, pp. 30--45, 1986.

\bibitem{9211490}
S.~D. Campbell, R.~P. Jenkins, P.~J. O'Connor, and D.~Werner, ``The explosion of artificial intelligence in antennas and propagation: How deep learning is advancing our state of the art,'' \emph{IEEE Antennas and Propagation Magazine}, vol.~63, no.~3, pp. 16--27, 2021.

\bibitem{baldi2021deep}
P.~Baldi, \emph{Deep learning in science}.\hskip 1em plus 0.5em minus 0.4em\relax Cambridge University Press, 2021.

\bibitem{482029}
Y.~Rahmat-Samii, L.~Williams, and R.~Yaccarino, ``The ucla bi-polar planar-near-field antenna-measurement and diagnostics range,'' \emph{IEEE Antennas and Propagation Magazine}, vol.~37, no.~6, pp. 16--35, 1995.

\bibitem{hertel2020sherpa}
L.~Hertel, J.~Collado, P.~Sadowski, J.~Ott, and P.~Baldi, ``Sherpa: Robust hyperparameter optimization for machine learning,'' \emph{SoftwareX}, 2020, in press.

\end{thebibliography}
\vfill

\end{document}